\documentclass[10pt, conference, letterpaper]{IEEEtran}

\usepackage{color,verbatim}
\usepackage{graphicx}
\usepackage{amsfonts}
\usepackage{amsmath}
\usepackage{mathtools}
\usepackage{amssymb}
\usepackage{bm}
\usepackage{color,verbatim}
\usepackage{multirow}
\usepackage{accents}
\usepackage{theoremref}
\usepackage{tabularx}
\usepackage{algorithmic}
\usepackage{algorithm}
\usepackage{url}
\usepackage{multicol}
\usepackage{array}
\usepackage[normalem]{ulem}
\usepackage{amsthm}
\usepackage{enumitem}
\usepackage[hidelinks]{hyperref}

\newtheoremstyle{paperstyle} % Name
  {5pt} % Space above
  {5pt} % Space below
  {} % Body font
  {} % Indent amount
  {\bfseries} % Theorem head font
  {:} % Punctuation after theorem head
  { } % Space after theorem head, ' ', or \newline
  {\thmname{#1}\thmnumber{ #2}\thmnote{ (#3)}} % Theorem head spec (can be left empty, meaning `normal')

\theoremstyle{paperstyle}
\newtheorem{theorem}{Theorem}

\IEEEoverridecommandlockouts

\begin{document}

    % HEADLINE
    \title{An Online Multiple Kernel \\ Parallelizable Learning Scheme\thanks{This work was supported by the SFI Offshore Mechatronics grant 237896/O30 and the PETROMAKS Smart-Rig grant 244205/E30 from the Research Council of Norway.}}
    \author{
    \IEEEauthorblockN{Emilio Ruiz-Moreno\IEEEauthorrefmark{1} and Baltasar Beferull-Lozano\IEEEauthorrefmark{1}\IEEEauthorrefmark{2},~\IEEEmembership{Senior Member,~IEEE}} \\
    \vspace{-5pt}
    \IEEEauthorblockA{\IEEEauthorrefmark{1}WISENET Center, Department of ICT, University of Agder, Grimstad, Norway} \\
    \vspace{-10pt}
    \IEEEauthorblockA{\IEEEauthorrefmark{2}SIGIPRO Department, Simula Metropolitan Center for Digital Engineering, Oslo, Norway}
    }
    \maketitle
    
    % SECTIONS
    %%% ABSTRACT
\begin{abstract}
The performance of reproducing kernel Hilbert space-based methods is known to be sensitive to the choice of the reproducing kernel.
Choosing an adequate reproducing kernel can be challenging and computationally demanding, especially in data-rich tasks without prior information about the solution domain.
In this paper, we propose a learning scheme that scalably combines several single kernel-based online methods to reduce the kernel-selection bias.
The proposed learning scheme applies to any task formulated as a regularized empirical risk minimization convex problem.
More specifically, our learning scheme is based on a multi-kernel learning formulation that can be applied to widen any single-kernel solution space, thus increasing the possibility of finding higher-performance solutions. 
In addition, it is parallelizable, allowing for the distribution of the computational load across different computing units.
We show experimentally that the proposed learning scheme outperforms the combined single-kernel online methods separately in terms of the cumulative regularized least squares cost metric.
\end{abstract}


    % INTRODUCTION 
\section{Introduction}
Reproducing kernel Hilbert space (RKHS)-based methods allow modeling highly non-linear relationships at a moderate computational cost \cite{scholkopf2002learning}.
Thanks to their simplicity and generality, they have been successfully adopted in a wide range of signal-processing applications \cite{perez2004kernel,takeda2007kernel}.

The performance of any RKHS-based method strongly relies on a preselected reproducing kernel (RK).
The efficient selection of an adequate RK presumes some task-specific prior information, such as knowledge about the data domain, invariant data transformations, geometrical data structures, or some properties of the underlying data generating process \cite{pozdnoukhov2006prior}.
For example, spline interpolation RKs are best suited for smooth data \cite{wahba1990spline,nosedal2012reproducing}.
Similarly, radial basis RKs can perform poorly if their associated hyperparameters are not properly tuned to the task.
The kernel-selection issue cannot be easily mitigated via cross-validation \cite{stone1974cross} because the associated computational load grows prohibitively with the number of RKs.
On the other hand, efficiently computable approximations of the leave-one-out error \cite{chapelle2002choosing} or hyperparameter optimization techniques \cite{williams2006gaussian} usually involve non-convexity and may lead to undesirable local minima.

Multi-kernel methods compensate for the lack of task-specific prior information using a predefined set of RKs known as \emph{dictionary}.
The dictionary can be formed by integrating different types of RKs, the same RK with different hyperparameter values, or a mix of both.
Typically, the preselected RK is constructed as a combination of several RKs from the dictionary \cite{gonen2011multiple}.
Therefore, how the dictionary is formed and how the preselected RK is constructed have a pivotal impact on the resulting accuracy and complexity of the method.
For instance, the larger the dictionary is, the more likely it is to reduce the kernel-selection bias compared to a particular RK or hyperparameter choice.
In addition, larger dictionaries allow for greater adaptability when learning from data samples since, in practice, these samples may come from a combination of different sources.
On the other hand, increasing the dictionary size becomes computationally demanding or even prohibitive.
For this reason, a commonly sought goal for multi-kernel methods is to find a compromise between performance and a computationally light and compact representation of the proposed solution in terms of the dictionary elements \cite{orabona2011ultra}.

Another related scaling issue that also applies to single-kernel methods is the curse of kernelization, i.e., potentially unbounded linear growth in model size with the amount of data \cite{wang2012breaking}.
This drawback is generally addressed through online approaches, which may rely on sparsification procedures \cite{tropp2004greed,engel2004kernel,richard2008online,souza2017fixed,souza2022online} or dimensionality reduction approximations \cite{rahimi2007random,money2023sparse}, among others \cite{dekel2005forgetron}.

In this context, some works \cite{sahoo2014online,shen2018online,sahoo2019large} have explored the use of online methods for determining the optimal solution associated with each single RK within the dictionary, as well as the best combination of these single-kernel solutions under a given task.
Following a conceptually similar approach, this paper proposes a novel multi-kernel learning scheme that can be parallelized across RKs by efficiently combining the solution of several single RK-based online methods running concurrently. 
This provides scalability with respect to the number of data samples and adaptability across different data patterns.
Moreover, it allows to distribute the computational load across different computing units as the dictionary size increases.
Our proposed scheme applies to any task that can be formulated as a regularized empirical risk minimization (RERM) convex problem \cite{vapnik1991principles,marteau2019beyond}.
Finally, the performance of the proposed learning scheme is experimentally validated in terms of the cumulative regularized least squares cost metric.

    % PROBLEM FORMULATION
\section{Problem formulation}
Supervised learning is arguably one of the core topics in machine learning \cite{shalev2014understanding}.
Many supervised learning tasks can be formulated as RERM convex problems whose solution admits a kernel representation.
That is, given a set of $N$ data samples $\mathcal{S} = \{(x^{(n)},y^{(n)})\}^N_{n=1}\subseteq\mathcal{X}\times\mathcal{Y}$, and an RKHS\footnote{The notation $\mathcal{Y}^\mathcal{X}$ refers to the set of functions from $\mathcal{X}$ to $\mathcal{Y}$.} $\mathcal{H}\subseteq\mathcal{Y}^\mathcal{X}$, the goal is to find a function estimate $f\in\mathcal{H}$ minimizing the following regularized functional cost
\begin{equation} \label{eq:functional_cost}
    \mathcal{C}_{\eta}\left(f;\mathcal{S}\right) = \sum^N_{n=1} \ell\left( f( x^{(n)}) , y^{(n)} \right) + \frac{\eta}{2} \Vert f \Vert^2_{\mathcal{H}} ,
\end{equation}
where the loss $\ell:\mathcal{Y}^2\to\mathbb{R}\cup\{\infty\}$ is a proper convex function used as a goodness-of-fit metric, the regularizer $\Vert \cdot \Vert^2_\mathcal{H}:\mathcal{H}\to\mathbb{R}$ is the squared RKHS $\mathcal{H}$ induced norm, and the hyperparameter $\eta\in\mathbb{R}_+$ controls the model complexity of the solution.

Under a multi-kernel learning framework, one typically constructs a valid RK by adequately combining the RKs within a preselected dictionary \cite{gonen2011multiple}.
Particularly, finding the RK within a convex hull of $P$ positive definite RKs that yields the function estimate incurring the lowest functional cost \eqref{eq:functional_cost} is equivalent to obtaining a solution from $\mathcal{H}$ built as the RKHS direct sum $\mathcal{H}_1\oplus\dots\oplus\mathcal{H}_P$, where each \textit{p}th RKHS $\mathcal{H}_p = \overline{\text{span}}\{k_p(x,\cdot):x\in\mathcal{X}\}$, being $k_p:\mathcal{X}^2\to\mathbb{R}$ its associated RK \cite{micchelli2005learning}.
The solution $f\in\mathcal{H}$ that minimizes \eqref{eq:functional_cost}, can be expressed without loss of generality as $f = \bm{\theta}^\top\bm{f}$, where $\bm{f}=[f_1,\dots,f_P]^\top\in\mathcal{H}_{1:P}$, with $\mathcal{H}_{1:P}\triangleq\mathcal{H}_1\times\cdots\times\mathcal{H}_P$ and $\bm{\theta}=[\theta_1,\dots,\theta_P]^\top\in\Delta^{P-1}$, with $\Delta^{P-1}\triangleq\{\bm{\beta}\in\mathbb{R}^P:\bm{\beta}\succeq\bm{0} \text{ and } \bm{1}^\top\bm{\beta} = 1\}$ denoting a simplex \cite{shen2019random}.
Thus, the RERM problem posed before becomes 
\begin{subequations} \label{eq:mkl_problem}
\begin{align}
    \underset{\bm{\theta}\in\Delta^{P-1},\bm{f}\in\mathcal{H}_{1:P}}{\text{ min }}& \mathcal{C}_\eta \left( f ; \mathcal{S} \right) \label{seq:mkl_problem_objective} \\
    \text{subject to: }& f = \bm{\theta}^\top\bm{f} .
\end{align}
\end{subequations}

Optimization problem \eqref{eq:mkl_problem} is bi-convex, meaning that it is convex in $\bm{\theta}$ for a fixed $\bm{f}$ and vice-versa. 
Still, it is not jointly convex in both optimization variables. 
It can be tackled via specialized methods that primarily exploit the convex substructures of the problem \cite{gorski2007biconvex}. 
However, these methods do not scale well with the number of RKs and data samples, denoted as $P$ and $N$, respectively.

This paper presents a method to solve efficiently \eqref{eq:mkl_problem} for large $P$ and $N$.

    % PROPOSED SOLUTION
\section{Proposed solution}
This section describes how to synergize an online formulation and an upper bound on the objective \eqref{seq:mkl_problem_objective} to solve \eqref{eq:mkl_problem} scalably with respect to $N$ and $P$.

%% ONLINE SETTING
\subsection{Online setting} \label{ssec:online_setting}
\begin{figure}
    \centering
    \includegraphics[width=0.95\columnwidth]{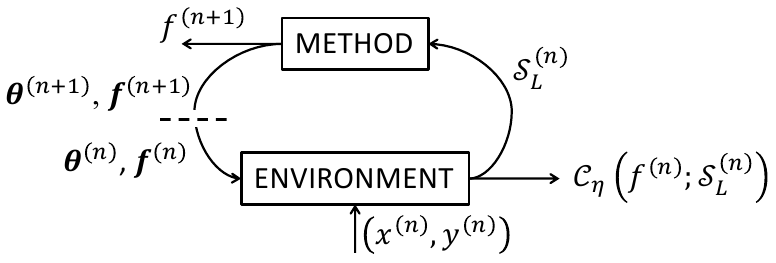}
    \caption{Visual description of the online setting considered.}
    \label{fig:online_setting}
\end{figure}
Online settings \cite{hazan2016introduction,orabona2019modern} can be adopted to solve \eqref{eq:mkl_problem} achieving low run-time complexity with respect to $N$ while incurring a certain tolerable (cumulative) cost.
They usually trade-off solution accuracy for speed, e.g., by processing only a few samples every iteration, for low memory complexity, e.g., by discarding samples after a few processing steps, or for model complexity control, i.e., bounded model size regardless of whether $N$ increases.

The online setting considered in this work can be cast as a method-environment iterative game \cite{zinkevich2003online}. 
The data samples in $\mathcal{S}$ are assumed to be available sequentially.
Then, at each iteration step $n$, the method chooses a function estimate $f^{(n)}\in\mathcal{H}_1\oplus\dots\oplus\mathcal{H}_P$ expressible as $f^{(n)} = {\bm{\theta}^{(n)}}^\top\bm{f}^{(n)}$ with $\bm{\theta}^{(n)}\in\Delta^{P-1}$ and $\bm{f}^{(n)}\in\mathcal{H}_{1:P}$.
In response, the environment penalizes the proposed function estimate $f^{(n)}$ with the incurred cost $\mathcal{C}_\eta(f^{(n)};\mathcal{S}^{(n)}_L)$, where $\mathcal{S}^{(n)}_L=\{(x^{(i)},y^{(i)})\}^n_{i=n_L}\subseteq\mathcal{S}$ is a sliding window of $L$ data samples and $n_L \triangleq \text{max}\{n-L+1,1\}$.
Finally, once the \textit{n}th function estimate $f^{(n)}$ is chosen, the method receives the \textit{n}th data window $\mathcal{S}^{(n)}_L$, which can be used at the next iteration\footnote{At the first iteration step $n=1$, the method has not received any data sample, thus $f^{(1)}$ is set as some arbitrary initial function estimate.} step $n+1$.
Fig. \ref{fig:online_setting} visually describes the procedure. 
 
%% PARALLELIZABLE UPPER BOUND
\subsection{Upper bound on the functional cost}
As we show next, we can improve scalability with respect to the number of RKs, by making use of the following upper bound:
\begin{subequations} \label{eq:upper_bound}
\begin{align}
    \mathcal{C}_\eta(f;\mathcal{S}) &= \sum^N_{n=1}\ell\left(\bm{\theta}^\top\bm{f}(x^{(n)}),y^{(n)}\right) + \frac{\eta}{2}\Vert\bm{\theta}^\top\bm{f}\Vert^2_\mathcal{H} \\
    &\leq \sum^N_{n=1}\sum^P_{p=1}\theta_p\ell\left(f_p(x^{(n)}),y^{(n)}\right) + \theta^2_p\frac{\eta}{2}\Vert f_p\Vert^2_{\mathcal{H}_p} \label{seq:upper_bound_step} \\
    &\triangleq \breve{\mathcal{C}}_\eta(\bm{\theta},\bm{f};\mathcal{S}) . \label{seq:upper_bound_cost}
\end{align}
\end{subequations}
The first upper-bounded term in \eqref{seq:upper_bound_step} follows directly from Jensen's inequality \cite{boyd2004convex}, whereas the second term is obtained by invoking the definition of the RKHS direct sum norm \cite{aronszajn1950theory}.
Even though the second term in \eqref{seq:upper_bound_step} could have also been upper bounded through Jensen's inequality, as in \cite{shen2019random}, exploiting the definition of the RKHS direct sum norm, leads to a tighter bound because $\theta_p^2 \leq \theta_p$ for $\theta_p \in [0,1]$.

The key advantage of the upper bound cost \eqref{seq:upper_bound_cost} is that it is separable across the $P$ RKs within the dictionary, hence allowing for parallelization at the expense of some loss in optimality, albeit with still satisfactory performance. 

%% LEARNING SCHEME
\subsection{Parallelizable learning scheme} \label{ssec:learning_scheme}
% Pseudo-code
\begin{algorithm}[t]
    \caption{(Quadratic program) Projection onto simplex.}
    \label{alg:projection_onto_simplex}
    \begin{algorithmic}[1]
        \renewcommand{\algorithmicrequire}{\textbf{Input:}}
        \renewcommand{\algorithmicensure}{\textbf{Output:}}
        % INPUT
        \REQUIRE The function estimate components $\{f_p^{(n)}\}^P_{p=1}$, the data window $\mathcal{S}^{(n-1)}_L$ and the loss $\ell$.
        % INITIALIZE
        \STATE Set the components of $\bm{a}$ and $\bm{b}$ as in \eqref{eq:components}.
        % ALGORITHM
        \STATE Sort $\bm{b}$ in ascending order, denoted as $\bm{u}$: $u_1 \leq u_2 \leq \dots \leq u_P$ and rearrange $\bm{a}$ accordingly into $\bm{v}$.
        \STATE Find $\rho = \text{max} \left \{1\leq j \leq P : u_j - \frac{\sum^j_{i=1} \frac{u_i}{v_i} + 2}{\sum^j_{i=1} \frac{1}{v_i}} < 0 \right \}$. \label{lagl:rho}
        \STATE Define $\mu = - \frac{2 + \sum^\rho_{i=1} \frac{u_i}{v_i}}{\sum^\rho_{i=1}\frac{1}{v_i}}$.
        \STATE Compute $\theta_p = \text{max}\left\{-\frac{1}{2a_p}(b_p + \mu) , 0 \right\}$ for $p=1,\dots,P$.
        \ENSURE $\bm{\theta}$.
    \end{algorithmic} 
\end{algorithm}

Our proposed learning scheme consists of executing at each iteration step $n$ the following consecutive operations:

1) Every \textit{p}th function estimate component $f^{(n)}_p$ is chosen through a single-kernel online method operating over the $p$th RK within the dictionary.
The methods are selected by the user, and they can be different for each RK as long as all of them adopt the online setting described in Sec. \ref{ssec:online_setting}.
For example, stochastic gradient descent methods for function estimation \cite{kivinen2004online}, and associated variants \cite{koppel2020optimally,ruiz2021tracking}, can be readily used.
Since the function estimate components of $\bm{f}^{(n)}$ can be computed in parallel across $P$ different computing units, the computational cost can be distributed.

2) Next, the convex weights in $\bm{\theta}^{(n)}$ are chosen as the ones that minimize the partially evaluated upper bound cost \eqref{seq:upper_bound_cost} at $\bm{f}^{(n)}$ and $\mathcal{S}^{(n-1)}_L$, referred from now on as learning cost.
Mathematically, 
\begin{subequations} \label{eq:quadratic_problem}
\begin{align}
    \bm{\theta}^{(n)} &= \text{ arg } \underaccent{\bm{\theta}\in\Delta^{P-1}}{\text{ min }}\,\,\, \breve{\mathcal{C}}_\eta \left( \bm{\theta} , \bm{f}^{(n)} ; \mathcal{S}^{(n-1)}_L \right) \\
    &= \text{ arg }\underaccent{\bm{\theta}\in\Delta^{P-1}}{\text{ min }}\,\,\, \bm{\theta}^\top \bm{A}^{(n)} \bm{\theta} + {\bm{b}^{(n)}}^\top\bm{\theta} , \label{seq:quadratic_problem_explicit}
\end{align}
\end{subequations}
where $\bm{A}^{(n)} \triangleq \text{diag}(\bm{a}^{(n)})\in\bm{S}^{P}_{++}$, with $\bm{a}^{(n)}\in\mathbb{R}^P_+$, and $\bm{b}^{(n)}\in\mathbb{R}^P$ whose components are computed\footnote{Notice that nothing prevents $f^{(n)}_p$ to be zero-valued and thus $\bm{A}^{(n)}$ from being singular and positive semi-definite. However, we can always set $\bm{A}^{(n)} = \text{diag}(\bm{a}^{(n)}) + \delta\bm{I}_P$ where $\delta$ is an arbitrarily small positive value.} as
\begin{subequations} \label{eq:components}
\begin{align}
    a^{(n)}_p &= \frac{\eta}{2} \left\Vert f^{(n)}_p \right\Vert^2_{\mathcal{H}_p} , \label{seq:a_component}\\
    b^{(n)}_p &= \sum_{i\in\mathcal{I}^{(n-1)}_L} \ell \left( f^{(n)}_p (x^{(i)}),y^{(i)} \right) \label{seq:b_component} ,
\end{align}
\end{subequations}
for all $p\in\mathbb{N}^{[1,P]}$, where $\mathcal{I}^{(n-1)}_L$ corresponds to the index set associated with the data samples in $\mathcal{S}^{(n-1)}_L$.
We adapt the projection onto the simplex algorithm discussed in \cite{shalev2006efficient,wang2013projection,blondel2014large} by extending its applicability to any quadratic problem described by a diagonal positive definite matrix with simplex constraints.
As a result, the proposed \textbf{Algorithm \ref{alg:projection_onto_simplex}} can solve \eqref{eq:quadratic_problem} exactly.
Its computational complexity is bottlenecked by a sorting step; that is, an asymptotic average complexity $\mathcal{O}(P\text{log}P)$ \cite{mishra2008selection}.
It should be mentioned that this complexity can be further reduced to $\mathcal{O}(P)$ on average by using a randomized pivot algorithm variation that identifies the parameter $\rho$ (\textbf{Algorithm \ref{alg:projection_onto_simplex}}, line \ref{lagl:rho}) using a divide and conquer procedure instead of sorting \cite{duchi2008efficient}, but this is out of the scope of the present paper.

3) Finally, the function estimate $f^{(n)} = {\bm{\theta}^{(n)}}^\top\bm{f}^{(n)}$ is proposed.

In summary, our scheme can be seen as a higher-level learner that iteratively chooses the lowest incurring learning cost combination of function estimates provided by lower-level learners, namely, single RK methods.
    % PERFORMANCE ANALYSIS
\section{Performance analysis} \label{sec:performance}
Under an online setting, as the one described in Sec. \ref{ssec:online_setting}, the incurred cost accumulated over time receives the name of \emph{cumulative} cost (CC).
In our case, the CC up to the \textit{n}th time step is given by $\sum^n_{i=1}\mathcal{C}_\eta(f^{(i)};\mathcal{S}^{(i)}_L)$.
From here, recall that every \textit{i}th function estimate $f^{(i)}$ is proposed via Sec. \ref{ssec:learning_scheme}, before the \textit{i}th data window $\mathcal{S}^{(i)}_L$ becomes available; thus, the CC is a measure of performance protecting against overfitting.
Intuitively, the lower the growth of the incurred CC with respect to $n$, the better the expected performance over unseen data.
In fact, popular measures of performance, such as the dynamic regret, are constructed as the difference between the CC incurred by the sequence of function estimates proposed by a method and a sequence of comparators \cite{jadbabaie2015online}.

In order to validate our scheme experimentally, we pose a signal reconstruction online problem from synthetically generated streaming data.
Specifically, we use the squared loss, i.e., $\ell(f(x),y) = (f(x)-y)^2$, and a dictionary of $P=20$ Gaussian kernels, i.e., $k_p(x,t) = \text{exp}\left( -\frac{1}{2}(x-t)^2/\sigma_p^2\right)$, with different widths $\sigma_p$ linearly spaced between $0.1$ and $10$. 
The method associated with each RK is an augmented naive online $R_\text{reg}$ minimization algorithm (NORMA) \cite{kivinen2004online} with a window length of $L=10$ data samples, a budget of $\tau = 100$ kernel expansion terms (beyond the allowed budget we truncate the oldest terms of the kernel expansion), and a fixed learning rate $\lambda_{\text{\tiny NORMA}}=0.05$. 
That is, before any possible truncation, each \textit{n}th function estimate associated with the \textit{p}th RK is constructed as $f^{(n)}_p = \sum^{n-1}_{i=1} \alpha^{(n)}_{p,i} \, k_p(x^{(i)},\cdot)$, where each $\alpha^{(n)}_{p,i}\in\mathbb{R}$ denotes a kernel-expansion coefficient obtained from the following NORMA update:
\begin{equation}
\label{eq:norma_update}
    f^{(n)}_p = f^{(n-1)}_p - \left. \lambda_{\text{\tiny NORMA}} \,  \partial_f \, \mathcal{C}_\eta \left(f;\mathcal{S}^{(n-1)}_L \right) \right\vert_{f=f^{(n-1)}_p} ,
\end{equation}
which, after some algebraic steps, leads to the next closed-form update rule:
\begin{equation}
\label{eq:norma_update_cf}
\alpha^{(n)}_i = 
\begin{cases}
    - \lambda_{\text{\tiny NORMA}} \, \ell'^{(n-1)}_{p,i} & \text{ if } i = n-1, \\
    \gamma\alpha^{(n-1)}_i - \lambda_{\text{\tiny NORMA}} \, \ell'^{(n-1)}_{p,i} & \text{ if }  i \in \mathcal{I}^{(n-1)}_L \backslash \{n-1\}, \\
    \gamma\alpha^{(n-1)}_i & \text{otherwise},
\end{cases}
\end{equation}
where $\ell'^{(n)}_{p,i} \triangleq \ell' \left( f^{(n)}_p(x^{(i)}),y^{(i)} \right) = 2\left( f^{(n)}(x^{(i)}) - y^{(i)} \right)$ and $\gamma \triangleq (1 - \lambda_{\text{\tiny NORMA}}\eta) \in \mathbb{R}^{(0,1)}_+$.
The regularization parameter is chosen as $\eta=0.01$. Lastly, the data samples have been generated via a stable AR(1) process $y^{(n)} = \varphi y^{(n-1)} + u^{(n)}$, with $\varphi=0.5488135$, $u^{(n)}\overset{\text{iid}}{\sim} \mathcal{N}(0,0.71519837)$, $y^{(0)} = 0$ and unit time stamps uniformly arranged in time, i.e., $x^{(n)}=n$.

Additionally, we compare our scheme with the online multiple kernel regression (OMKR) algorithm \cite{sahoo2014online}, arguably the closest approach conceptually. 
More specifically, we compare against the budget OMKR gradient-based variant method over the same experimental setting described above.
Briefly, the considered OMKR method can be described, at each iteration step $n$, by the following three-stage scheme:

1) The set of function estimates, in this case, regressors proposed by each one of the $P$ NORMAs, is updated as \eqref{eq:norma_update_cf} and collected in $\bm{f}^{(n)} \in \mathcal{H}_{1:P}$.

2) Then, the $P$ weights for combining the multiple regressors are updated as
\begin{equation}
\label{eq:update_omkr}
    \bm{w}^{(n)} = \bm{w}^{(n-1)} -  \lambda^{(n)}_\text{\tiny OMKR} \, \left. \nabla_{\bm{w}} \, \mathcal{C}_\eta \left( \bm{w}^\top \bm{f}^{(n)} ; \mathcal{S}^{(n-1)}_L \right) \right\vert_{\bm{w} = \bm{w}^{(n-1)}} .
\end{equation}
In this case, we use an initial learning rate $\lambda^{(1)}_\text{\tiny OMKR} = 8\cdot10^{-4}$ that is halved every 50 steps until a minimum value of $10^{-5}$.
After some algebraic manipulations and making use of the definition of the RKHS direct sum norm \cite{aronszajn1950theory}, the evaluated gradient in \eqref{eq:update_omkr} equals to
\begin{equation}
    \sum_{i\in\mathcal{I}^{(n-1)}_L} \hspace{-3mm} \bm{f}^{(n)}(x^{(i)})\ell' \left( {\bm{w}^{(n-1)}}^\top\bm{f}^{(n)}(x^{(i)}),y^{(i)} \right) + \eta \bm{A}^{(n)} \bm{w}^{(n-1)} .
\end{equation}
The matrix $\bm{A}^{(n)}$ corresponds to the one introduced in \eqref{seq:a_component}, and the initial combination weights are set as $\bm{w}^{(1)} = \bm{0}_P$.

3) Finally, the function estimate $f^{(n)} = {\bm{w}^{(n)}}^\top \bm{f}^{(n)}$ is proposed.

Unlike our scheme, the OMKR algorithm can eventually learn any linear combination of single-kernel function estimates.
However, due to the additive nature of the update step in \eqref{eq:update_omkr}, the OMKR algorithm usually suffers from slow convergence rates. 
Moreover, it requires a sequence of learning rates $\lambda^{(n)}_{\text{\tiny OMKR}}$ whose tuning involves optimization techniques or task-specific knowledge, hence adversely affecting performance, e.g., poor learning or instabilities, if not carried out adequately.

\begin{figure}[!t]
    \centering
    \includegraphics[width=0.95\columnwidth]{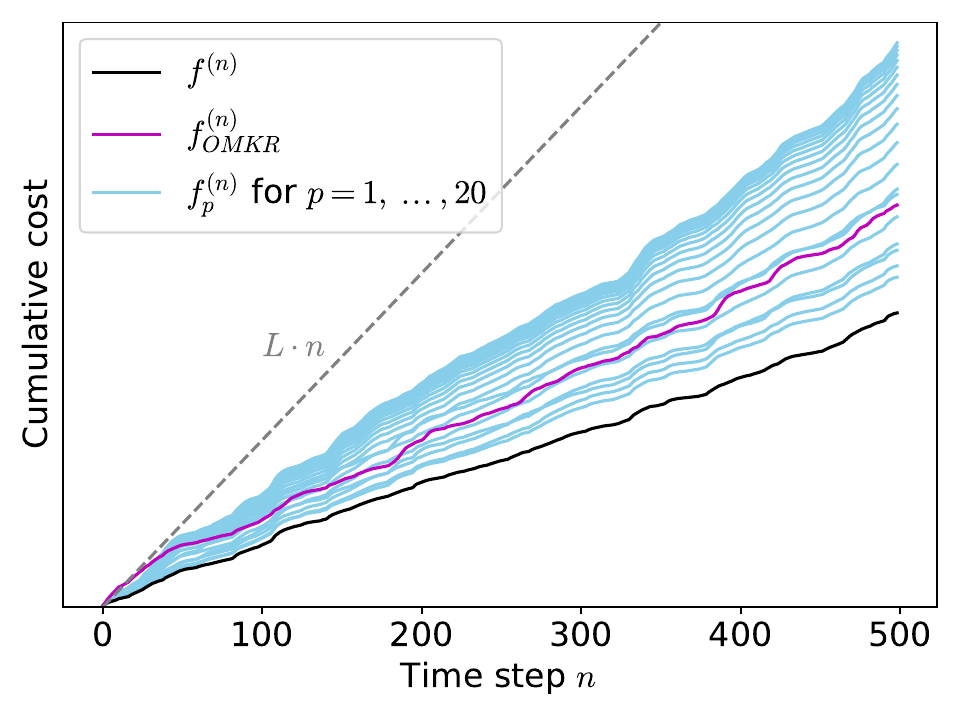}
    \caption{Cumulative cost up to time step $n$ incurred by our learning scheme, the OMKR algorithm, and the combined single RK NORMA regressors individually. The dashed-dotted line $L\cdot n$ is shown as a reference.}
    \label{fig:cumulative_cost}
\end{figure}

\begin{figure}[!t]
    \centering
    \includegraphics[width=0.95\columnwidth]{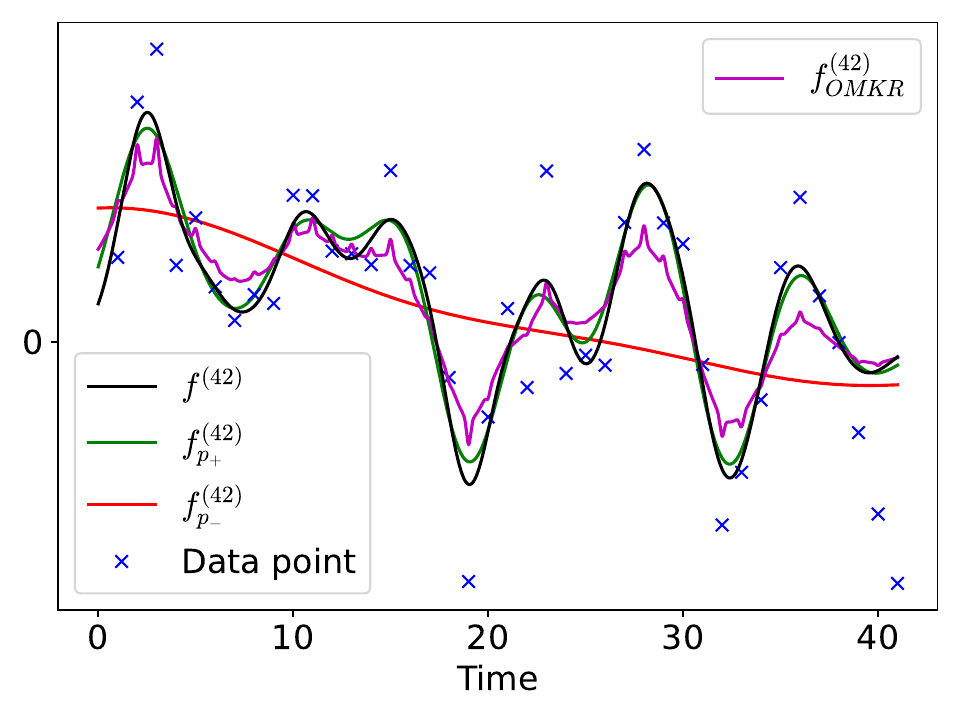}
    \caption{Snapshot of the $42$nd signal estimate obtained by the OMKR algorithm, our learning scheme, and the best and the worst of the combined single RK NORMA regressors (denoted by the indices $p_+$ and $p_-$, respectively) in terms of the so-far ($n=42$) incurred cumulative cost.}
    \label{fig:reconstruction}
\end{figure}

Our experimental results in Fig. \ref{fig:cumulative_cost} show that the CC incurred by our proposed learning scheme outperforms the lowest CC incurred by any of the combined single RK NORMA regressors separately.
In the same figure, it can also be observed that our scheme incurs a CC that increases at a lower rate than the one incurred by the OMKR algorithm, thus allowing our scheme to outperform the best NORMA regressor much sooner.
The reason behind this observation is arguably the additive nature of the OMKR algorithm, which requires numerous updates to completely remove the residual contributions of irrelevant regressors.
For example, see in Fig. \ref{fig:reconstruction} the ``spiky'' shape of the OMKR signal estimate due to some regressors constructed with a narrow-valued $\sigma_p$ RK.

Regarding computational resources, both the OMKR algorithm and our scheme can be parallelized across the RKs within the dictionary, which in our experiments means a constant time complexity $\mathcal{O}(\tau)$, and a combination step of complexity $\mathcal{O}(P)$ and $\mathcal{O}(P\,\text{log}P)$, respectively.
However, as mentioned in Sec. \ref{ssec:learning_scheme}, the complexity of our scheme can be further reduced to $\mathcal{O}(P)$ on average, making both of the compared approaches computationally equivalent.


    % CONCLUSION
\section{Conclusion}
We present a multi-kernel learning scheme that experimentally outperforms the best of the combined single RK methods, in terms of the cumulative regularized least squares cost metric, with a comparable computational load per computing unit.
This corroborates the ability of the proposed scheme to effectively accommodate a larger function space (from which to draw function estimates) of multi-kernel methods while keeping the lower computational complexity of online single RK methods. 
Furthermore, although \textbf{Algorithm \ref{alg:projection_onto_simplex}} has been expressly designed for the task discussed in this paper, it can be used to solve any other problem that accepts a formulation as in \eqref{seq:quadratic_problem_explicit}.

    % BIBLIOGRAPHY
    \bibliography{references}

    % SUPPLEMENTARY
    \newpage
\onecolumn
\begin{center}
\Huge
    Supplementary material for ``An Online Multiple Kernel Parallelizable Learning Scheme''
\end{center}
\setcounter{equation}{0}
\renewcommand{\thesection}{S.\Roman{section}} 
\renewcommand{\theequation}{S.\arabic{equation}}
\medskip

% SUPPLEMENTARY
\section*{Correctness of \textbf{Algorithm 1}} \label{sec:correctness}
A Lagrangian of problem (4) is 
\begin{equation}
    \text{L}(\bm{\theta}, \bm{\lambda}, \mu) = \bm{\theta}^\top \text{diag}(\bm{a}^{(n)})\bm{\theta} + {\bm{b}^{(n)}}^\top \bm{\theta} - \bm{\lambda}^\top\bm{\theta} + \mu(\bm{1}^\top\bm{\theta} - 1) ,
\end{equation}
being $\mu\in\mathbb{R}$ and $\bm{\lambda} \in \mathbb{R}^P$ the Lagrange multipliers associated with the equality and inequality constraints, respectively. 
At the optimal solution $\bm{\theta}^{(n)}$, the following KKT conditions \cite{nocedal1999numerical} hold:
\vspace{-1pt}
\begin{subequations} \label{eq:KKT_alg2}
    \begin{alignat}{2}
        2\theta^{(n)}_p a^{(n)}_{p} + b^{(n)}_{p} - \lambda_p + \mu &= 0, &&\hspace{15pt} p = 1,\dots,P \label{eq:stationarity} \\
        \theta^{(n)}_p &\geq 0, &&\hspace{15pt} p = 1, \dots, P \label{eq:primal_in_constraint} \\
        \lambda_p &\geq 0, &&\hspace{15pt} p = 1, \dots, P \label{eq:dual_constraint} \\
        \lambda_p \theta^{(n)}_p &= 0, &&\hspace{15pt} p = 1, \dots, P \label{eq:complementary_slackness}\\
        \sum^P_{p=1} \theta^{(n)}_p &= 1 . \label{eq:primal_eq_constraint}
    \end{alignat}
\end{subequations}
From the complementary slackness, stated in \eqref{eq:complementary_slackness}, we can deduce that if the primal inequality constraint in \eqref{eq:primal_in_constraint} is slacked, i.e., greater than zero, then $\lambda_p = 0$ and from the stationarity condition \eqref{eq:stationarity}, the solution fulfils 
\begin{equation} \label{eq:if_theta_slack}
    \theta^{(n)}_p = - \frac{1}{2 a^{(n)}_{p}} (b^{(n)}_{p} + \mu) > 0 .
\end{equation}
On the other hand, if the primal inequality constraint is tight, i.e., $\theta^{(n)}_p = 0$, then the dual constraint (\ref{eq:dual_constraint}) is not binding. 
Again, from the stationarity condition in \eqref{eq:stationarity}, we can identify those non-binding constraints as those that satisfy the following expression:
\begin{equation} \label{eq:not_binding}
    b^{(n)}_{p} + \mu = \lambda_p \geq 0 .
\end{equation}
In this way, it is clear from \eqref{eq:not_binding} that the components of the optimal solution that are zero, if any, correspond to the larger components of $\bm{b}^{(n)}$.
Without loss of generality, we can assume that the components of $\bm{b}^{(n)}$ are sorted in ascending order as long as the components of $\bm{a}^{(n)}$ are rearranged accordingly.
Thus, by comparing $\bm{b}^{(n)}$ with the solution as follows:
\begin{equation}
    \begin{split}
        b^{(n)}_{1} \leq \cdots \leq b^{(n)}_{\rho} \leq \, & b^{(n)}_{\rho+1} \leq \cdots \leq b^{(n)}_{P}, \\
        & \theta^{(n)}_{\rho+1} = \cdots = \theta^{(n)}_P = 0 ,
    \end{split}
\end{equation}
it can be concluded that the index $\rho \in \mathbb{N}^{[1,P]}$ determines the number of components in the solution that are nonzero. 
From here, and rewriting the equality primal constraint (\ref{eq:primal_eq_constraint}) as
\begin{equation}
    \sum^P_{p=1} \theta^{(n)}_p = \sum^\rho_{p=1} \theta^{(n)}_p = - \frac{1}{2} \sum^\rho_{p=1} \frac{1}{a^{(n)}_{p}}(b^{(n)}_{p} + \mu) = 1 ,
\end{equation}
the Lagrangian multiplier associated with the equality constraint can be isolated and computed as
\begin{equation} \label{eq:mu}
    \mu = - \frac{2 + \sum^\rho_{p=1} \frac{b^{(n)}_{p}}{a^{(n)}_{p}}}{\sum^\rho_{p=1} \frac{1}{a^{(n)}_{p}}} ,
\end{equation}
as long as the index $\rho$ is known.

%% THEOREM (how to obtain rho)
\begin{theorem} \label{thm:rho}
Let $\rho$ be the number of positive components in the solution of optimization problem (4), then
\begin{equation} \label{eq:rho}
    \rho = \text{max} \left \{ 1 \leq j \leq P : b_j - \frac{2 + \sum^j_{i=1} \frac{b_i}{a_i}}{\sum^j_{i=1}\frac{1}{a_i} } < 0 \right \} ,
\end{equation}
where $\bm{b}$ is obtained by sorting $\bm{b}^{(n)}$ components in ascending order and $\bm{a}$ corresponds to $\bm{a}^{(n)}$ rearranged accordingly.
\end{theorem}

%% proof
\textit{Proof}.
Let us first define the quantities $\varphi_j \triangleq b_j - (2 + \sum^j_{i=1}\frac{b_i}{a_i})/\sum^j_{i=1}\frac{1}{a_i}$ and $s_{j:k} \triangleq \sum^k_{i=j} \frac{1}{a_i}$.
Then, the goal is to show that $j=\rho$ is the largest index in $\{1,\dots,P\}$ for which $\varphi_j$ remains negative.

For $j<\rho$, we have that
\begin{subequations} \label{eq:j<r}
\begin{align}
    % eq line 1
    \varphi_j &= \frac{1}{s_{1:j}}\left(s_{1:j} b_j - \left(2 + \sum^j_{i=1}\frac{b_i}{a_i}\right)\right) \label{seq:j<r_1} \\
    % eq line 2
    &= \frac{1}{s_{1:j}} \left( s_{1:j} b_j -2 - \sum^\rho_{i=1} \frac{b_i}{a_i} + \sum^\rho_{i=j+1} \frac{b_i}{a_i} \right) \label{seq:j<r_2} \\
    % eq line 3
    &= \frac{1}{s_{1:j}} \left( s_{1:j} b_j + s_{1:\rho} \mu + \sum^\rho_{i=j+1} \frac{b_i}{a_i} \right) \label{seq:j<r_3} \\
    % eq line 4
    &= b_j + \mu + \frac{1}{s_{1:j}} \sum^\rho_{i=j+1} \frac{1}{a_i} (\mu + b_i) < 0 \label{seq:j<r_6} ,
\end{align}
\end{subequations}
where in step \eqref{seq:j<r_2} we use the equivalence $\sum^j_{i=1} \frac{b_i}{a_i}=\sum^\rho_{i=1} \frac{b_i}{a_i}-\sum^\rho_{i=j+1} \frac{b_i}{a_i}$. 
Next, in step (\ref{seq:j<r_3}), we make use of the relation in \eqref{eq:mu}. 
Finally, the step \eqref{seq:j<r_6} holds thanks to the relation in (\ref{eq:if_theta_slack}) and because $s_{1:j},a_i \geq 0$ $\forall i,j$. 

For $j=\rho$, and thanks to \eqref{eq:if_theta_slack}, we have $\varphi_\rho = b_\rho + \mu < 0$.
Then, using the relation in \eqref{eq:mu}, we can verify that \eqref{eq:rho} holds.

For $j>\rho$, we can follow similar algebraic steps as in \eqref{eq:j<r} to obtain
\begin{equation} \label{eq:j>r}
     \varphi_j = \frac{1}{s_{1:j}} \left( s_{1:\rho} (b_j + \mu) + \sum^j_{i=\rho+1}\frac{1}{a_i}(b_j - b_i) \right) \geq 0 .
\end{equation}
The inequality in \eqref{eq:j>r} holds thanks to the relation in (\ref{eq:not_binding}) and the fact that $b_j \geq b_i$ $\forall i$.
    
\end{document}